\documentclass[11pt,a4paper]{article}
\usepackage[hyperref]{acl2020}
\usepackage{times}
\usepackage{latexsym}

\usepackage{microtype}

\usepackage[linesnumbered,ruled,vlined]{algorithm2e}
\usepackage{bm}
\usepackage{amsmath}
\usepackage{mathtools}

\usepackage{booktabs}
\usepackage{amsfonts}
\usepackage{xcolor}
\usepackage{multirow}

\aclfinalcopy 

\usepackage{array}
\newcommand{\PreserveBackslash}[1]{\let\temp=\\#1\let\\=\temp}
\newcolumntype{C}[1]{>{\PreserveBackslash\centering}p{#1}}
\newcolumntype{R}[1]{>{\PreserveBackslash\raggedleft}p{#1}}
\newcolumntype{L}[1]{>{\PreserveBackslash\raggedright}p{#1}}

\title{Frustratingly Hard Evidence Retrieval for QA Over Books}

\author{
    Xiangyang Mou \\
    Rensselaer Polytechnic Institute \\
    Troy, NY 12180 \\
    \texttt{moux4@rpi.edu}
    \\\And
    Mo Yu \\
    IBM Research  \\
    USA  \\
    \texttt{yum@us.ibm.com}
    \\\And
    Bingsheng Yao \\
    Rensselaer Polytechnic Institute \\
    Troy, NY 12180 \\
    \texttt{yaob@rpi.edu}
    \\\AND
    Chenghao Yang \\
    Columbia University \\
    New York, NY 10027 \\
    \texttt{chenghao.yang@columbia.edu}
    \\\And
    Xiaoxiao Guo \\
    IBM Research  \\
    USA  \\
    \texttt{xiaoxiao.guo@ibm.com}
    \\\AND
    Saloni Potdar \\
    IBM Watson  \\
    USA  \\
    \texttt{potdars@us.ibm.com}
    \\\And
    Hui Su \\
    IBM Research \\
    USA \\
    \texttt{huisuibmres@us.ibm.com}
}

\date{}

\begin{document}
\maketitle

\begin{abstract}

A lot of progress has been made to improve question answering (QA) in recent years, but the special problem of QA over narrative book stories has not been explored in-depth. We formulate BookQA as an open-domain QA task given its similar dependency on evidence retrieval. We further investigate how state-of-the-art open-domain QA approaches can help BookQA. Besides achieving state-of-the-art on the NarrativeQA benchmark, our study also reveals the difficulty of evidence retrieval in books with a wealth of experiments and analysis - which necessitates future effort on novel solutions for evidence retrieval in BookQA.


\end{abstract}


\section{Introduction}

The task of question answering has benefited largely from the advancements in deep learning, especially from the pre-trained language models(LM)~\cite{radford2019language,devlin2018bert}. While question answering over single passage (reading comprehension datasets) and over the large-scale open-domain corpora~(open-domain QA) have largely benefited from these, the performance of QA over book stories (BookQA) lags behind. For example, the most representative benchmark in this direction, the NarrativeQA~\cite{kovcisky2018narrativeqa} which was released three years ago - the current state-of-the-art methods only show marginal improvement over the first baselines.

There are several challenges in NarrativeQA which slow down the research progress. First, the narrative stories lead to a new writing style which differs from previous works over formal texts like Wikipedia. Second, the long inputs of books are beyond the processing ability of neural models so evidence identification from a whole book is critical. Third, NarrativeQA is a generative task, and many of the answers cannot be exactly matched in the original books. Hence, the generative QA models are required. Finally and most importantly, the dataset does not provide annotations of the supporting evidence. While this makes it a realistic setting like open-domain QA, together with the generative nature of the answers, also makes it difficult to infer the supporting evidence similar to most of the extractive open-domain QA tasks.

The requirements around evidence identification and the missing supporting evidence annotation make BookQA task similar to open-domain QA. In this paper, we first study whether the ideas used in state-of-the-art open-domain QA systems can be extended to improve BookQA including: (1) the neural ranker-reader pipeline~\cite{wang2018r}, where a neural ranker is used to select related passages (evidence) given a question from a large candidate sets; (2) the usage of pre-trained LMs as reader and ranker, such as GPT~\cite{radford2019language}, BERT~\cite{devlin2018bert} and their follow-up work; (3) the distantly supervised and unsupervised training techniques~\cite{wang2018r,Lee_2019orqa,min2019discrete,guu2020realm,karpukhin2020dense} that help rankers learn more from noisy gold data.


By training a ranker-reader framework on BookQA, we successfully achieve a new state-of-the-art on NarrativeQA using both generative and extractive readers. Based on these results and our analysis, we observe the followings: 

\noindent$\bullet$ Using the pre-trained LMs as the reader model, such as BERT and GPT, improves the NarrativeQA performance. With the same BM25 IR baseline, they give 5-6\% improvement on Rouge-L over their non-pre-trained counterparts.

\noindent$\bullet$ Our specifically designed distant supervision signals improve the neural ranker significantly, but the improvement is small compared to the upper bound. Further analysis of the ranker module confirms the difficulty in training, as the improvement from the pre-trained LM BERT is marginal in it.

\section{Proposed Method}

\subsection{Task Definition}
Following~\cite{kovcisky2018narrativeqa}, we define the task of \textbf{BookQA} as finding the answer $\mathbf{A}$ to a question $\mathbf{Q}$ from a book $\mathbf{B}$,\footnote{To be more accurate, the question should be denoted as $\mathbf{Q_B}$ but we use $\mathbf{Q}$ for simplicity.} where each book contains a number of consecutive paragraphs $\mathcal{C}$ (usually hundreds or more). $\mathbf{A}$ is a free-form answer that can be concluded from the book but may not appear in it in an exact form.

In this paper we propose an open-domain QA formulation and solution to the task of BookQA. Specifically, the task consists of (1) an evidence retrieval step that selects evidence from $\mathbf{B}$ for $\mathbf{Q}$, which in our case is a collection of paragraphs $\mathcal{C}_{\mathbf{Q}}=\{\mathbf{C_i}\}\subset\mathbf{B}$; and (2) a question-answering step that predicts an answer given $\mathbf{Q}$ and $\mathcal{C}_{\mathbf{Q}}$.

In the state-of-the-art open-domain QA systems, the aforementioned two steps are modeled by two learnable models (usually based on pre-trained LMs), namely the \textbf{ranker} and the \textbf{reader}.
The ranker predicts the relevance of each paragraph $\mathbf{C} \in \mathbf{B}$ to the question, where the top ranked paragraphs form the $\mathcal{C}_{\mathbf{Q}}$; and the reader predicts the answer following
$P(\mathbf{A}|\mathbf{Q}, \mathcal{C}_{\mathbf{Q}})$.

In the following subsections, we describe our solution to make the training of pre-trained LM-based ranker and reader work for the BookQA task.

\subsection{Reader (QA Model)}
\label{ssec:model_qa}

\paragraph{Extractive Reader}
We use a pre-trained BERT model \cite{devlin2018bert, Wolf2019HuggingFacesTS} to predict the answer span given the query and the context. One challenge of training an extraction model in BookQA is that there is no annotation of true spans because of its generative nature. Our solution is to find the most likely span as answer supervision. Specifically, we compute the Rouge-L score \cite{lin2004rougeL} between the true answer and each candidate span of the same length, and finally take the span with the maximum Rouge-L score as our weak label. We initially tried the exact-answer spans but failed to find many due to its low coverage in BookQA.


\paragraph{Generative Reader}
We try GPT-2~\cite{radford2019language} and BART~\cite{lewis2019bart} pre-trianed LMs for answer generation. By design, both GPT-2 and BART are autoregressive models and therefore do not require additional annotations for training. Considering the memory limit, we use the \texttt{GPT-2-medium}  and \texttt{bart-large} as our pre-trained generative models and fine-tune it on BookQA mostly using default training parameters\footnote{https://huggingface.co/transformers/model\_doc/gpt2.html}. If the output contains several sentences, we only choose the first one.


\subsection{Book Paragraph Ranker}
\label{ssec:model_ranker}

We fine-tune another BERT binary classifier for paragraph retrieval, following the usage of BERT on text similarity tasks. In BookQA, training such a classifier is challenging because of the lack of evidence-level supervision. We deal with this problem by using an ensemble method to achieve distant supervision. We build two weak BM25 retrievers with one using only $\mathbf{Q}$ and the other using both $\mathbf{Q}$ and true $\mathbf{A}$. Denoting the correspondent rough-grained retrievals as $\mathcal{C}_\mathbf{Q}$ and $\mathcal{C}_{\mathbf{Q}+\mathbf{A}}$, we then tutor a model to select their intersection $\mathcal{C}_\mathbf{Q} \cap \mathcal{C}_{\mathbf{Q}+\mathbf{A}}$ by sampling the positive samples from $\mathcal{C}_\mathbf{Q} \cap \mathcal{C}_{\mathbf{Q}+\mathbf{A}}$ and the negative ones from $(\mathcal{C}_\mathbf{Q} \cap \mathcal{C}_{\mathbf{Q}+\mathbf{A}})^{\mathsf{c}}$. In order to encourage the ranker to select passages that have better coverage of the answers, we further apply a \textbf{Rouge-L filter} upon the previous sampling results, and only select the positive samples whose answer-related Rouge-L score is higher than the upper threshold and the negative samples lower than the lower threshold\footnote{In practice, we set the hyperparameters 0.7 and 0.4}.

\begin{table*}[t!]
    \small
    \centering
    \begin{tabular}{lccc} 
        \toprule
        \bf System & \bf w/ trained ranker & \bf w/ pre-trained LM & \bf w/ extra training data \\
        \midrule
        \textcolor{red}{Attention Sum} \cite{kovcisky2018narrativeqa}& &\\
        \textcolor{blue}{BiDAF} \cite{kovcisky2018narrativeqa} & &\\
        \textcolor{red}{IAL-CPG} \cite{tay2019simple}&  & &    \\
        \textcolor{blue}{R$^3$} \cite{wang2017r3}& $\checkmark$ & & \\
        \textcolor{blue}{BERT-heur} \cite{frermann-2019-extractive} &  $\checkmark$ & $\checkmark$ & $\checkmark$\\
        Our \textcolor{red}{generative}/\textcolor{blue}{extractive} systems & $\checkmark$ & $\checkmark$ & \\
        \bottomrule
    \end{tabular}
    \caption{Summary of the characteristics of the compared systems. \textcolor{red}{Red}/\textcolor{blue}{blue} color refers to generative/extraction QA systems. In addition to the standard techniques, \cite{wang2017r3} uses reinforcement learning to train the ranker; and \cite{tay2019simple} uses curriculum to train the reader to overcome the divergence of evidence retrieval qualities between training and testing.}
    \vspace*{-1mm}
    \label{tab:model_cheatsheet}
\end{table*}


\section{Experiments}

\subsection{Settings}
\label{ssec:settings}

\paragraph{Dataset}

We conduct experiments on NarrativeQA dataset \cite{kovcisky2018narrativeqa}, which has a collection of 783 books and 789 movie scripts and their summaries, with each having on average 30 question-answer pairs. Each book or movie script contains an average of 62k words. NarrativeQA provides two different settings, the \textbf{summary} setting and the \textbf{full-story} setting. Our BookQA task corresponds to the full-story setting that finds answers from books or movie scripts. Note that the NarrativeQA is a \emph{generative} QA task. The answers are not guaranteed to appear in the books.

We preprocess the raw data with SpaCy\footnote{\url{https://spacy.io/}} tokenization. Then following~\cite{kovcisky2018narrativeqa}, we cut the books into non-overlapping paragraphs with a length of 200 each for the full-story setting.

\paragraph{Baseline}
We conduct experiments with both generative and extractive readers, and compare with the competitive baseline models from \cite{kovcisky2018narrativeqa, tay2019simple, frermann-2019-extractive} in the full-story setting. Meanwhile, we take a BM25 retrieval as the baseline ranker and evaluate our distantly supervised BERT rankers. We also compare to the strong results from~\cite{frermann-2019-extractive}, which constructed evidence-level supervision with the usage of book summaries. However, the summary is not considered available by design~\cite{kovcisky2018narrativeqa} in the general full-story scenario where questions should be answered solely from books.\footnote{In NarrativeQA, the summary has a good coverage of the answers due to the data collection procedures; also, summaries can be viewed as humans' comprehension of the books.}


Although not the focus of the paper, our reader performance in the summary setting is also reported (Section~\ref{ssec:exp_reader}), to show the properties of the readers.

\paragraph{Metrics}
Because of the generative nature of the task, following previous works~\cite{kovcisky2018narrativeqa,tay2019simple,frermann-2019-extractive}, we evaluate the QA performance with Bleu-1, Bleu-4~\cite{papineni2002bleu}, Meteor~\cite{banerjee2005meteor}, Rouge-L~\cite{lin2004rougeL}.\footnote{We used an open-source evaluation library~\cite{sharma2017nlgeval}: \url{https://github.com/Maluuba/nlg-eval}.}
We also report the Exact Match(EM) and F1 scores\footnote{The squad/evaluate-v1.1.py script is used.} that are commonly used in open-domain QA evaluation. We convert both hypothesis and reference to lowercase and remove the punctuation before evaluation. 

\paragraph{Model Selection} We select the best models on the development set according to its average score of Rouge-L and EM. For ranker model selection, we use the average score of upper bound EM and Rouge-L of top-5 ranked paragraphs.




\subsection{Reader Model Validation (the~QA-over-Summary~Setting)}
\label{ssec:exp_reader}

First, we compare our readers under the summary setting, to verify the correctness of our readers.
Our BERT reader achieves performance close to the public state-of-the-art in this setting.

Our GPT-2 reader outperforms the existing systems without usage of pointer generators (PG), but is behind the state-of-the-art with PG. 
Despite the large gap between systems with and without PG in this setting, \cite{tay2019simple} demonstrates that it didn't contribute much in the full-story setting in the ablation study. Nonetheless, we will investigate the usage of PG in pre-trained LMs in the future work.



\begin{table*}[t!]
    \small
    \centering
    \begin{tabular}{lcccccc} 
        \toprule
        \bf System & \bf Bleu-1 & \bf Bleu-4 & \bf Meteor & \bf Rouge-L \\
        \midrule
        \multicolumn{5}{c}{\bf Extractive Readers}    \\
        BERT + Hard EM~\cite{min2019discrete}& - & - & - & \bf 58.1/58.8 \\
        BERT-only~\cite{min2019discrete}& - & - & - & 55.8/56.1  \\
        BERT w/ full training signals [\textbf{Ours}] &  49.35/49.02 & 25.76/25.85 & 23.93/24.14 & 52.62/52.02    \\
        BERT w/ exact answer match only [\textbf{Ours}] &  \bf49.78/49.64 & \bf27.01/28.94 & \bf 25.22/25.12 & 57.19/56.35  \\
        \midrule
        \multicolumn{5}{c}{\bf Generative Readers}    \\
        Attention Sum~\cite{kovcisky2018narrativeqa} (w/o PG) & 23.54/23.20 & 5.90/6.39 & 8.02/7.77 & 23.28/22.26   \\
        Masque~\cite{nishida2019multi} (w/ PG)&-/48.70 & -/20.98 & -/21.95 & -/54.74  \\
        GPT-2 Reader(w/o PG) [\textbf{Ours}] & 33.63/35.49 & 11.87/14.33 & 13.71/14.36 & 34.32/35.65  \\
        BART Reader(w/o PG) [\textbf{Ours}] & \textbf{57.22}/\textbf{56.20} & \textbf{28.78}/\textbf{29.41} & \textbf{27.41}/\textbf{26.60} & \textbf{60.46}/\textbf{58.76}  \\
        \bottomrule
    \end{tabular}
    \caption{Results under NarrativeQA summary setting on dev/test set (\%). PG refers to the usage of pointer generator. For extractive model, we compare with the best public result~\cite{min2019discrete} and its BERT-only ablation. The latter corresponds to the same setting as ours. For generative model, we compare with the best public models with and without pointer generators.
    }
    \vspace*{3mm}
    \label{tab:model_performance}
\end{table*}

\subsection{Main Results (the QA-over-Book Setting)}

We then experimented our whole QA pipelines in the full-story setting. Table~\ref{tab:generative_reader_model_performance} and Table~\ref{tab:extractive_reader_model_performance} compare our results with public state-of-the-art generative and extractive QA systems.

Our pipeline system with the baseline BM25 ranker outperforms the existing state-of-the-art, confirming the advantage of pre-trained LMs as observed in most QA tasks. Our distantly supervised ranker adds another 1-2\% of improvement to all the metrics, bringing both our generative and extractive models with the best performance.
It also helps outperform~\cite{frermann-2019-extractive} on multiple metrics without the usage of the strong extra supervision from the summaries.

\begin{table*}[t!]
    \small
    \centering
    \setlength{\tabcolsep}{5pt} 
    \renewcommand{\arraystretch}{1} 
    \begin{tabular}{lcccccc} 
        \toprule
        \bf System & \bf Bleu-1 & \bf Bleu-4 & \bf Meteor & \bf Rouge-L & \bf EM & \bf F1 \\
        \midrule
        & \multicolumn{3}{c}{\bf Public Generative Baselines}    \\
        AttSum (top-10)~\cite{kovcisky2018narrativeqa} & 20.00/19.09 & 2.23/1.81 & 4.45/4.29 & 14.47/14.03  & - & -    \\
        AttSum (top-20)~\cite{kovcisky2018narrativeqa} & 19.79/19.06 & 1.79/2.11 & 4.60/4.37 & 14.86/14.02 & - & -    \\
        
        IAL-CPG \cite{tay2019simple} &23.31/22.92 & 2.70/2.47 & 5.68/5.59 & 17.33/17.67 & - & -\\
        \quad - curriculum &20.75/- & 1.52/- & 4.65/-& 15.42/- \\
        \midrule
        & \multicolumn{3}{c}{\bf Our Generative QA Models}               \\
        BM25 + GPT-2 Reader      & 24.54/24.43 & 4.74/4.37 & 7.32/7.32 & 20.25/21.04  & 5.12/5.22 & 17.72/18.38     \\
        \quad + BERT Ranker      & 24.94/25.03 & 4.76/4.42 & 7.74/7.81 & 21.89/22.36  & 6.79/6.31 & 19.67/19.94    \\
        \quad + \it Oracle IR (BM25 w/ Q+A)   & \it 33.18/32.95 & \it 8.16/7.70 & \it 12.35/12.47 & \it 34.83/34.96 & \it 17.09/15.98 & \it 33.65/33.75\\ 
        
        BM25 + BART Reader      & 26.20/\textbf{26.69} & \textbf{4.95}/5.07 & 8.38/8.56 & 23.41/24.15  & 6.61/7.10 & 21.33/21.85    \\
        \quad + BERT Ranker     & \textbf{26.50}/26.62 & 4.79/\textbf{5.12} & \textbf{8.41/8.53} & \textbf{23.49/24.28}  & \textbf{6.85/7.34} & \textbf{21.34/22.10}    \\
        \quad + \it Oracle IR (BM25 w/ Q+A)   & \textit{37.66/37.43} & \textit{10.16/9.57} & \textit{14.66/14.81} & \textit{38.70/39.41} & \textit{18.28/18.19} & \textit{38.51/38.98}     \\ 
        \bottomrule
    \end{tabular}
    \caption{Generative performance in NarrativeQA full-story setting (BookQA setting) dev/test set(\%). \textit{Oracle IR} utilizes question and true answers for retrieval.}
    \label{tab:generative_reader_model_performance}
\end{table*}

\begin{table*}[t!]
    \small
    \centering
    \setlength{\tabcolsep}{5pt} 
    \renewcommand{\arraystretch}{1} 
    \begin{tabular}{lcccccc} 
        \toprule
        \bf System & \bf Bleu-1 & \bf Bleu-4 & \bf Meteor & \bf Rouge-L & \bf EM & \bf F1 \\
        \midrule
        \multicolumn{7}{c}{\bf Public Extractive Baselines}    \\
        BiDAF~\cite{kovcisky2018narrativeqa}              & 5.82/5.68 & 0.22/0.25 & 3.84/3.72 & 6.33/6.22 & - & -    \\
        R$^3$ \cite{wang2017r3} &\bf16.40/15.70 &0.50/0.49 &3.52/3.47 &11.40/11.90 & -&- \\
        \midrule
        \multicolumn{7}{c}{\bf Our Extractive QA Models}\\
        BM25 + BERT Reader     & 13.27/13.84 & 0.94/1.07 & 4.29/4.59 & 12.59/13.81 & 4.67/5.26 & 11.57/12.55    \\
        \quad + BERT Ranker   & 14.60/14.46 &  \textbf{1.81}/\textbf{1.38} & \textbf{5.09}/\textbf{5.03} & \bf 14.76/15.49 & \bf 6.79/6.66 &\bf 13.75/14.45 \\
        \quad + \it Oracle IR (BM25 w/ Q+A)&\it 23.81/24.01  & \it 3.54/4.01& \it 9.72/9.83&\it 28.33/28.72&\it 15.27/15.39&\it 28.42/28.55 \\
        \midrule
        \multicolumn{7}{c}{\bf Extractive Models w/ additional supervision}   \\
        BERT-heur \cite{frermann-2019-extractive} &  -/12.26& -/\textbf{2.06} &-/\textbf{5.28} & -/15.15 & - &-\\
        \bottomrule
    \end{tabular}
    \caption{Extractive performance in NarrativeQA full-story setting (BookQA setting) dev/test set(\%). \textit{Oracle IR} utilizes question and true answers for retrieval.}
    \label{tab:extractive_reader_model_performance}
\end{table*}

\begin{table*}[t!]
    \small
    \centering
    \setlength{\tabcolsep}{7pt} 
    \renewcommand{\arraystretch}{1} 
    \begin{tabular}{
        C{0.21\textwidth}
        C{0.21\textwidth}
        C{0.21\textwidth}
        C{0.21\textwidth}
        }
        \toprule
        \textbf{Question} & \textbf{Gold Answer 1} & \textbf{Gold Answer 2} & \textbf{Generative Result} \\
        \midrule
        Where is Millicent sent to boarding school? &
        Millicent is sent to a boarding school in France &
        France &
        France \\
        \midrule
        What is Morgan's relationship to Wyatt? &
        Morgan is Wyatt's brother   &
        Brothers    &
        Brother \\
        \midrule
        What illness does Doc Holiday suffer from? &
        Tuberculosis    &
        Tuberculosis    &
        Lung cancer     \\
        \midrule
        How does Carl make his house fly?   &
        He attaches thousands of helium balloons to it &
        Balloons &
        He uses a parachute to climb up the side of the dirigible   \\
        \midrule
        How does Felipe die?    &
        Suicide     &
        He suffers a physical breakdown     &
        He is killed by a bullet in the head    \\
        \midrule
        What was the great stone face and how did it appear?    &
        A natural rock formation on the side of a mountain &
        A natural rock formation which appeared when viewed at a proper distance     &
        It was a stone face \\

        \bottomrule
    \end{tabular}
    \caption{Generative result examples. The model tends to generate shorter answers in general. The longer answer it generates, the less likely the answer tends to be correct. The grammatical correctness and fluency of the long generative answers are approaching to human level, regardless of the problematic logic between the generated answer and question. The majority of the generative results do not make sense logically which leads to the low scores in different metrics.}
    \label{tab:generative_result_comparison}
\end{table*}

\subsection{Ablation of Ranker Performance}

To take a deeper look at the challenges in ranker training, we conduct an ablation study on the ranker independently. The quality of a ranker is measured by the answer coverage of its top-5 selections on the basis of the top-32 candidates from the baseline. The answer coverage is estimated by the maximum Rouge-L score of the subsequences of the selected paragraphs of the same length as the answers; and whether the answer can be covered by any of the selected paragraphs (EM).

Our BERT ranker together with supervision filtering strategy has a significant improvement over the BM25 baseline. Our BERT ranker improves by 0.7\%, compared with MatchLSTM~\cite{wang2016learning} or an improved BiDAF architecture~\cite{clark2018simple}. On the other hand, comparing the benefits that BERT brings to open-domain QA tasks, the relatively small improvement demonstrates the difficulty of evidence retrieval in BookQA. This shows the potential room for future novel improvements, which is also exhibited by the large gap between our best rankers and either the upper bound or the oracle.





\subsection{Discussion of Future Improvement}

We can see a considerable gap between our best models (ranker and readers) and their corresponding oracles in Table~\ref{tab:generative_reader_model_performance}, ~\ref{tab:extractive_reader_model_performance}, and~\ref{tab:ir_model_performance}. 
One difficulty that limits the effectiveness of ranker training is the noisy annotation resulted from the nature of the free-form answers. Our filtering technique helps significantly but is still not sufficient. One way we believe that can improve the distant supervision signals is by iteratively updating the ranker and reader like in Hard-EM~\cite{min2019discrete, guu2020realm}. Another possible direction is to extend the idea of inferring evidence on training data with game-theoretic approaches~\cite{perez2019finding,feng2020learning}, then use the inferred evidence paragraph as labels to train the ranker.

\begin{table}[t!]
    \small
    \centering
    \begin{tabular}{lcc}
        \toprule
        \bf IR Method & \bf EM & \bf Rouge-L         \\
        \midrule
        BM25 &18.99 & 47.48 \\
        BERT ranker  & \bf 24.26 & \bf 52.68 \\
        \quad - Rouge-L filtering & 22.63 & 51.02 \\
        \quad Repl BERT w/ BiDAF & 21.88 & 50.64 \\
        \quad Repl BERT w/ MatchLSTM & 21.97 & 50.39 \\
        \midrule
        Upperbound (BM25 top-32) & 30.81 & 61.40\\
        Oracle (BM25 w/ Q+A) & 35.75 & 63.92\\
        \bottomrule
    \end{tabular}
    \caption{IR Evaluation on NarrativeQA dev set(\%).}
    \vspace*{-2mm}
    \label{tab:ir_model_performance}
\end{table}


\section{Conclusion}
We explored the BookQA task and systemically tested on NarrativeQA dataset different types of models and techniques from open-domain QA. Our proposed approaches bring significant improvements to the state-of-the-art across different metrics. Our insight and analysis lay the path for exciting future work in this domain. 

\section*{Acknowledgment}
This work is supported by Cognitive and Immersive Systems Lab (CISL), a collaboration between IBM and RPI, and also a center in IBM’s AI Horizons Network.

\bibliography{reference}
\bibliographystyle{acl_natbib}

\end{document}